\documentclass{article}

\usepackage[nonatbib,preprint]{neurips_2025}

\usepackage[utf8]{inputenc} %
\usepackage[T1]{fontenc}    %
\usepackage{hyperref}       %
\usepackage{url}            %
\usepackage{booktabs}       %
\usepackage{amsfonts}       %
\usepackage{nicefrac}       %
\usepackage{microtype}      %
\usepackage{xcolor}         %

\usepackage{multirow}
\usepackage{graphicx}
\newcommand{\eg}{\textit{e.g.}}
\newcommand{\ie}{\textit{i.e.}}
\usepackage{amsmath}

\title{LiftVSR: Lifting Image Diffusion to Video Super-Resolution via Hybrid Temporal Modeling 
with Only 4$\times$RTX 4090s}

\vspace{-5pt}
\author{
  \vspace{-25pt}\\
  \textbf{Xijun Wang$^{1}$,\quad Xin Li$^{1}$\thanks{Corresponding Author (xin.li@ustc.edu.cn)},\quad Bingcheng Li$^{1}$,\quad Zhibo Chen$^1$} \vspace{5pt}\\
  $^1$University of Science and Technology of China \vspace{3pt} \\
  Project Page:~\, \url{https://kopperx.github.io/projects/liftvsr}\vspace{5pt} \\
}

\begin{document}

\maketitle

\vspace{-24pt}
\begin{figure}[h] %
    \centering %
    \includegraphics[width=\textwidth]{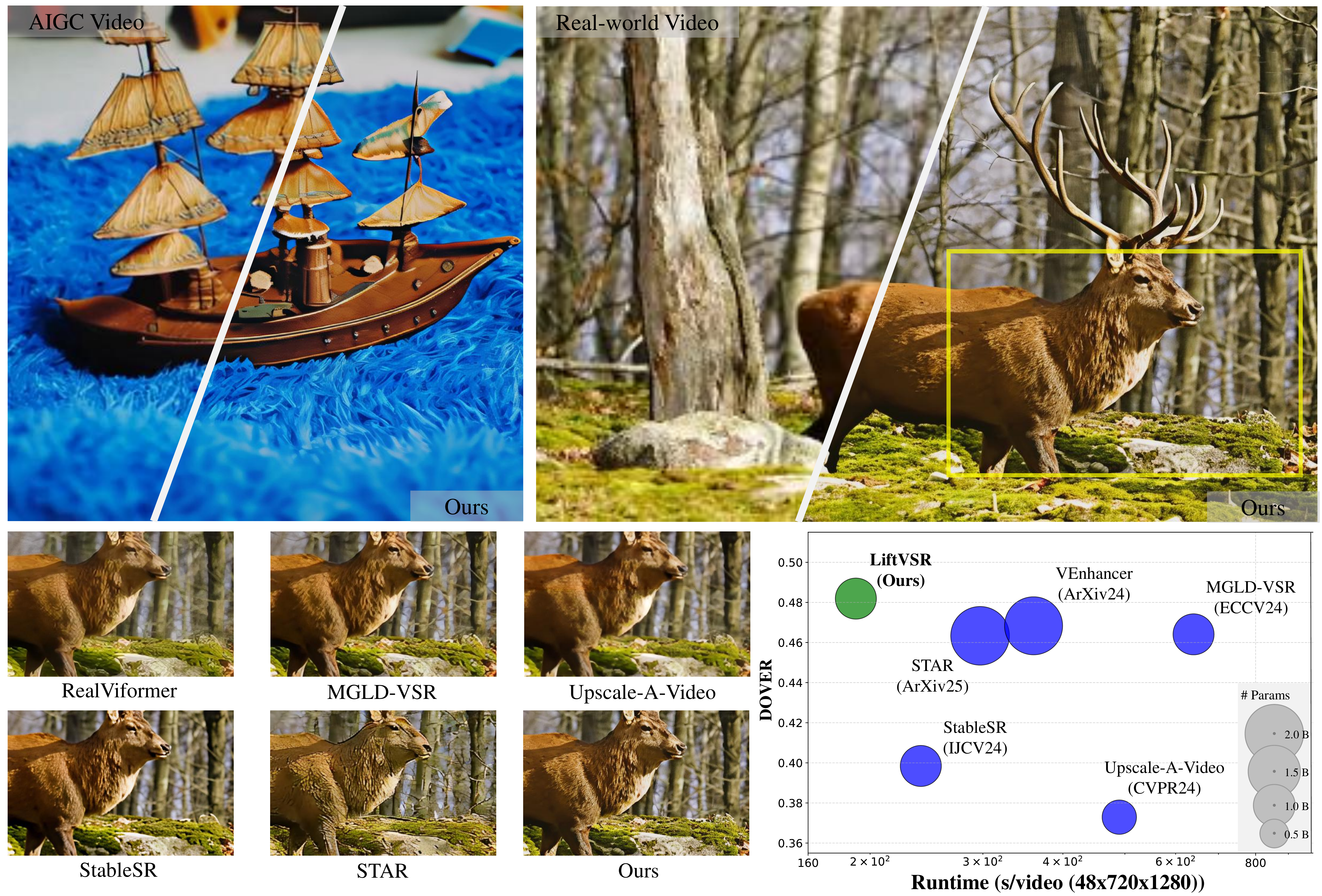}
    \caption{
        Visual and speed comparisons on both real-world and AI-generated videos.
        Our proposed LiftVSR achieves superior performance in terms of both visual quality and speed.
        \textbf{(Zoom-in for best view)}
    } %
    \label{fig:teaser} %
\end{figure}

\begin{abstract}
    
Diffusion models have significantly advanced video super-resolution (VSR) by enhancing perceptual quality, largely through elaborately designed temporal modeling to ensure inter-frame consistency. However, existing methods usually suffer from limited temporal coherence and prohibitively high computational costs (e.g., typically requiring over 8 NVIDIA A100-80G GPUs), especially for long videos.
In this work, we propose LiftVSR, an efficient VSR framework that leverages and elevates the image-wise diffusion prior from PixArt-$\alpha$, achieving state-of-the-art results using only 4$\times$RTX 4090 GPUs. To balance long-term consistency and efficiency, we introduce a hybrid temporal modeling mechanism that decomposes temporal learning into two complementary components: (i) Dynamic Temporal Attention (DTA) for fine-grained temporal modeling within short frame segment (\ie, low complexity), and (ii) Attention Memory Cache (AMC) for long-term temporal modeling across segments (\ie, consistency).
Specifically, DTA identifies multiple token flows across frames within multi-head query and key tokens to warp inter-frame contexts in the value tokens. AMC adaptively aggregates historical segment information via a cache unit, ensuring long-term coherence with minimal overhead. To further stabilize the cache interaction during inference, we introduce an asymmetric sampling strategy that mitigates feature mismatches arising from different diffusion sampling steps. Extensive experiments on several typical VSR benchmarks have demonstrated that LiftVSR achieves impressive performance with significantly lower computational costs.

\end{abstract}

\section{Introduction}
\label{sec:intro}
Real-world Video Super Resolution (VSR) aims to super-resolve low-resolution videos suffering from unknown and complicated hybrid degradations, which has garnered significant attention due to its application potential across both academic and industry fields, \eg, streaming media, post-processing, or content creation. Existing VSR works~\cite{basicvsr++,realbasicvsr,jo2018deep,edvr,toflow,liu2022learning} primarily focus on temporal contextual knowledge excavation and utilization through temporal modeling, including  (i) explicit optical flow estimation~\cite{basicvsr,toflow,cao2021video}, (ii) implicit dynamic feature matching/warping~\cite{edvr,basicvsr++,wang2019deformable,tdan}, and (iii) temporal attention mechanism~\cite{vrt,liu2022learning}, resulting in impressive objective quality.

However, higher objective quality metrics do not necessarily correlate with better perceptual experience, highlighting the urgent need for perception-oriented VSR. Early works~\cite{realbasicvsr,realesrgan,animesr,realviformer} achieve this by employing generative adversarial networks (GANs) to enhance subjective quality, where the unstable adversarial training paradigm relies on the carefully designed optimization routes. More recently, diffusion models~\cite{ldm,ddpm,ddim} have demonstrated the strong capability to model the distribution of high-resolution, high-quality images, inspiring the integration of diffusion-based generative priors into the VSR framework to further improve subjective quality.

Recent diffusion-based VSR methods can be roughly divided into two types based on pre-trained diffusion models: (i) Image diffusion-based VSR methods~\cite{uav,diffvsr,mgldvsr} aim to extend the image-wise diffusion priors to VSR by incorporating and optimizing temporal modules, such as 3D convolutions, temporal attention, and training-free optical flow guidance~\cite{uav}. Although it is efficient, existing temporal modeling inevitably exhibits temporal flickering and inconsistent details, particularly for long videos; (ii) In contrast, video diffusion-based VSR methods \cite{venhancer,star,seedvr} directly reuse the spatial-temporal generative priors from pre-trained video diffusion models for VSR, where the improved temporal coherence comes at the cost of large model sizes and high computational costs. For instance, STAR and VEnhancer~\cite{star, venhancer} were optimized with over 8\textasciitilde16 NVIDIA A100 GPUs 
and have over 2B billion parameters as shown in Fig.~\ref{fig:teaser}.

In this work, we present LiftVSR, an efficient VSR framework
based on a image-wise diffusion model PixArt-$\alpha$ \cite{pixart-alpha},
achieving state-of-the-art performance using only 4$\times$RTX 4090 GPUs.
We propose a novel hybrid temporal modeling mechanism 
that combines two complementary component: fine-grained temporal modeling within short frame segment, ensuring low computational complexity,
and long-term modeling across long video segments, preserving long-term consistency.
This dual component strikes a balance between long-term coherence and computational efficiency.

Specifically, we introduce a Dynamic Temporal Attention (DTA) module 
to achieving fine-grained temporal modeling within a fixed frame segment.
Conventional temporal attention is designed to model tokens at the same spatial location and relies on intra-frame self-attention to implicitly capture information from different spatial positions.
It struggles to capture accurate temporal correspondences under limited video data and model capacity.
In contrast, DTA explicitly learns token correspondences with multi-head query and key tokens.
In each head, we predict a token flow map from reference frames to non-reference frames with a flow estimation network.
Then, we warp and aggregate aligned tokens through an attention operation.
Compared to deformable attention \cite{rvrt, de-detr}, which
predicts multiple candidate locations within non-reference frames,
DTA identifies single flow path for each attention head that significantly reduces the number of tokens in one attention operation, thus further reducing the computing complexity.
The multi-head mechanism further provides the dynamics of token
aggregation, leading to robust and efficient temporal information aggregation.

For long-term temporal modeling across segments,
temporal attention exist significantly increased computational cost when processing long videos.
To balance the computational cost and long-term coherence,
we propose a novel Attention Memory Cache (AMC) module to enable long-term feature propagation.
The AMC module maintains a memory cache unit and adaptively aggregates historical segment information to it
with two key operations: query and update.
The query operation adaptively retrieves previous frame features from the cache unit
while the update operation refreshes the memory cache with features from current frame segment.
Additionally, a simple gating unit is introduced to adaptively update the memory cache
based on current frame segment.

Inspired by \cite{df}, we further introduce an asymmetric sampling strategy
to stabilize the cache interaction during sampling.
By mitigating the feature mismatches arising from different diffusion sampling steps,
we can effectively enable flexible cache conditions across segments as shown in Fig.~\ref{fig:sampling}.
Specifically, we use the cache from the last denoising step as conditions for next segment generation
to avoid noise accumulation and reduce the GPU memory usage.

With the integration of hybrid temporal modeling mechanism,
along with the asymmetric sampling strategy,
we attain robust and efficient long-term temporal modeling ability and significantly improved training and inference efficiency.
As illustrated in Fig. \ref{fig:teaser}, our model achieves the best Dover score \cite{dover} and the lowest time cost
with only 4$\times$RTX 4090 GPUs.
Extensive experiments on several benchmarks demonstrate that our model achieves 
state-of-the-art subjective quality and temporal coherence, outperforming existing methods.

\section{Related Work}
\label{sec:relate}
\paragraph{Video Super-Resolution.}
As an extension of image super-resolution (SR), VSR is a challenging task due to the additional temporal consistency constraint.
Traditional VSR methods can be broadly categorized into two types:
sliding-window \cite{edvr,mucan,tdan,toflow} and recurrent-based \cite{basicvsr,realbasicvsr,sajjadi2018frame,huang2017video,rsdn} approaches.
Sliding-window methods aggregate temporal information from a local window of frames to super-resolve the center frame,
while recurrent methods propagate information frame by frame 
using recurrent neural networks (RNNs) to enhance all frames sequentially.
They typically using optical flow or deformable convolutions (DCNs)~\cite{dcn,tdan} to align the frames in the sequence.
To better align with the real-world scenarios,
RealBasicVSR~\cite{realbasicvsr} introduced a realistic degradation synthesis pipeline 
to generate synthetic LQ-HQ pairs for training.
Additionally, it incorporates a pre-cleaning module designed to remove distortions prior to super-resolution processing.
And RealViformer~\cite{realviformer} introduces a channel attention mechanism which is proven to be less sensitive to artifacts.
It further enhances performance by employing squeeze-excite mechanisms and covariance-based rescaling. 
Despite the significant advancements these methods have achieved, 
traditional VSR methods often struggle to produce high-quality detials due to the lack of generative prior.

\paragraph{Diffusion Prior for Video Task.}
Diffusion models have recently achieved remarkable success in various image-level tasks, 
including generation~\cite{ddpm,ldm}, editing~\cite{p2p,kawar2023imagic}, and enhancement\cite{stablesr,osediff,diffbir,diffir}, 
attracting widespread attention in the research community.
Naturally, there has been a growing interest in extending image diffusion models to video tasks \cite{vdm,videop2p,uav,qi2023fatezero}.
The core challenge in lifting image diffusion models to videos lies in temporal modeling.
Early works~\cite{tune-a-video,flatten,qi2023fatezero,yang2023rerender} attempted to address this in a zero-shot manner,
leveraging cross-frame attention or optical guidance to ensure temporal consistency.
More recently, to further improve the temporal coherence and generalization of output videos,
\cite{vdm,animatediff,i2vgen-xl,svd} proposed to pretrain diffusion models on large-scale video data.
Building upon image diffusion prior, 
these methods have made significant progress
by introducing temporal module like 3D convolution or temporal attention.
More recently, DiT-based video generation models ~\cite{cogvideox,latte,xu2024easyanimate} have emerged.
~\cite{cogvideox} incorporates 3D attention mechanisms to further improve performance.
However, these models typically involve huge computational costs, resulting in slow speed for both training and inference, 
which severely limits their real-world application.
In this work, we introduce a hybrid temporal modeling mechanism which aims to lift image diffusion models to real-world VSR tasks in a more efficient manner.

\paragraph{Diffusion Prior for Video Super-Resolution.}
With the rapid development of diffusion models,
there has been a growing interest in applying diffusion priors to real-world VSR tasks.
Several recent works~\cite{uav,seedvr,star,mgldvsr} 
have been proposed and have demonstrated strong performance.
We can categorize these methods into two main groups:
1) those that utilize pre-trained image diffusion models and extend them into 3D architectures,
and 2) those that directly leverage pre-trained video diffusion models.
For instance, Upscale-A-Video, MGLDVSR, DiffVSR, and SeedVR ~\cite{uav,mgldvsr,diffvsr,seedvr} etc.
leverage pre-trained Stable Diffusion \cite{ldm} and 
extend them into 3D architecture by incorporating temporal layers.
However, these methods often struggle to maintain temporal consistency 
due to the limited modeling capacity of their temporal components, especially in long videos.
In contrast, VEnhancer \cite{venhancer} and STAR \cite{star} directly utilize pre-trained video diffusion models, 
leading to better temporal consistency.
But these methods face unaffordable computational costs when video length increases,
resulting in limited temporal modeling scope.
In this paper, we explore how to efficiently lift a single-frame diffusion model to real-world VSR tasks,
and propose a novel DiT-based VSR pipeline, LiftVSR, equipped with hybrid temporal modeling mechanism,
achieving state-of-the-art performance with significantly lower training cost and higher inference speed.

\section{Methodology}
\label{sec:method}
\begin{figure}[h] %
    \centering %
    \includegraphics[width=\textwidth]{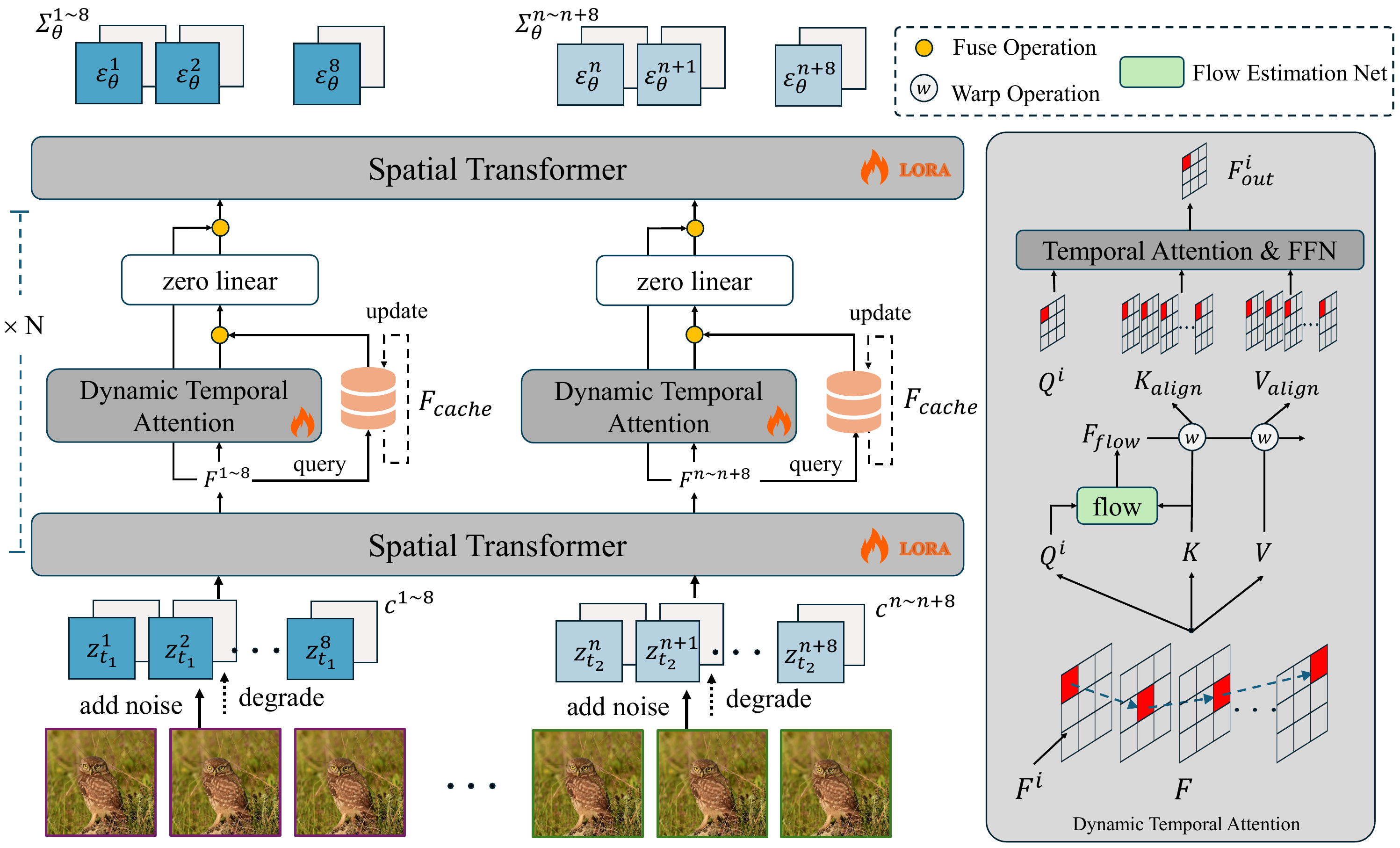}
    \caption{
        An overview of our LiftVSR.
        Our LiftVSR is built upon a pre-trained DiT model.
        It incorporates hybrid temporal modeling mechanisms to ensure long-range temporal consistency.
        Additionally, it employs asymmetric sampling strategy to support flexible cache interaction across segments.
        } 
    \label{fig:model} 
\end{figure}

Given a low-quality video, our goal is to enhance its resolution while maintaining temporal consistency and improving visual quality.  
In this paper, we propose a novel pipeline, LiftVSR, with hybrid temporal modeling mechanisms and asymmetric sampling strategy.
An overview of the proposed LiftVSR is shown in Fig. \ref{fig:model}.
The LiftVSR is built upon a pre-trained DiT model \cite{pixart-alpha}, 
including Dynamic Temporal Attention (DTA) module for fine-grained temporal modeling within short-term segments,
and Attention Memory Cache (AMC) module for long-term temporal feature propagation.
We further introduce an asymmetric sampling strategy to mitigate feature 
mismatches arising from different diffusion sampling steps to enable flexible cache interaction across segments
as shown in Fig. \ref{fig:sampling}.
With the proposed innovations, LiftVSR achieves significant improvements in temporal consistency 
and visual quality while maintaining low training and inference costs.

\subsection{Preliminary: Latent Diffusion Model}
Latent Diffusion Models (LDMs)~\cite{ldm} have emerged as a powerful generative modeling framework
by gradually transforming data into Gaussian noise and then learning to reverse it.
Given a sample $x_0$, LDMs first encode it into a latent representation $z_0$ using a VAE encoder $\mathcal{E}$.
Then, the diffused $z_t = \sqrt{\alpha_t} z_0 + \sqrt{1 - \alpha_t} \epsilon, \epsilon \sim \mathcal{N}(0, I)$ 
is generated by a forward diffusion process, 
where $\alpha_t$ is a predefined variance schedule and $\epsilon$ is random Gaussian noise.
The denoising network $\epsilon_\theta(z_t, t, c)$ is designed to predict the noise $\epsilon$ 
at each time step $t$ given the noisy latent $z_t$ and condition $c$.
The optimization objective can be formulated as follows:
\begin{equation}
    L = \mathbb{E}_{z, t, \epsilon, c} \left[ ||\epsilon - \epsilon_\theta(z_t, t, c)||^2 \right]
\end{equation}
where $c$ can be text, images or any other control signals.
During inference, the model generates samples by starting from pure noise $z_T$ and 
iteratively denoising it to obtain the final latent $z_0$.
The latent representation $z_0$ is then decoded back to the image space eith a VAE decoder $\mathcal{D}$.

\subsection{Dynamic Temporal Attention}
To enable temporal level modeling in the image diffusion model,
we introduce a Dynamic Temporal Attention (DTA) module to capture
fine-grained temporal context within a fixed frame segment.
In contrast to deformable attention~\cite{rvrt,de-detr}, which directly predicts multiple candidate locations in non-reference frames,
DTA identifies single flow path for each attention head and enables dynamic 
token aggregation through multi-head mechanism.
DTA reduces the number of tokens in one attention operation, thus significantly reducing the computing complexity.

Specifically, given a video feature $F \in \mathbb{R}^{n \times h \times w \times d}$, 
where $n$ is the number of frames, and $d$ is the dimension of the feature.
We first choose a reference frame $F^i \in \mathbb{R}^{h \times w \times d}$ from $F$,
where $i$ is the index of the reference frame.
For each head,
we can obtain the query, key and value tokens by linear projection:
$Q^i = F^i W_q, K = F W_k, V = F W_v$,
where $W_q, W_k, W_v \in \mathbb{R}^{d \times \frac{d}{nheads}}$ are the projection matrices.
Note that the query $Q^i$ is computed from the reference frame $F^i$.
We then concatenate the $Q^i$ and $K$ tokens and 
feed them into a flow estimation network to generate a token flow map $F_{flow} \in \mathbb{R}^{n \times h \times w \times 2}$.
The flow estimation network consists of a stack of convolutional layers and activation functions.
Then the generated flow map $F_{flow}$ is to warping $K$, $V$ tokens,
obtaining aligned feature token $K_{align}, V_{align} \in \mathbb{R}^{n \times h \times w \times \frac{d}{nheads}}$.
\begin{equation}
    K_{align} = \mathcal{W}(K, F_{flow}), V_{align} = \mathcal{W}(V, F_{flow}),
\end{equation}
where $\mathcal{W}()$ denotes flow warping operation.
Then we reshape the aligned feature to get $\hat{K}_{align} \in \mathbb{R}^{(hw) \times n \times \frac{d}{nheads}}$, $\hat{V}_{align} \in \mathbb{R}^{(hw) \times n \times \frac{d}{nheads}}$,
and apply the attention mechanism to obtain the aggregated feature $\hat{F}_{out}^i \in \mathbb{R}^{(hw) \times 1 \times \frac{d}{nheads}}$.
\begin{equation}
    \hat{F}_{out}^i = \mathrm{Attention}(Q^i, \hat{K}_{align}, \hat{V}_{align}) = softmax(\frac{Q^i\hat{K}_{align}^T}{\sqrt{d}}) \hat{V}_{align}
\end{equation}
Finally, we concatenate $\hat{F}_{out}^i$ over all heads and reshape it to obtain
 $F_{out}^i \in \mathbb{R}^{1 \times h \times w \times d}$ and 
apply FFN layer to get the final output.
By iterating all frames as reference frames, 
we can get the DTA output $F_{out} \in \mathbb{R}^{n \times h \times w \times d}$
finally.

\subsection{Attention Memory Cache}
Due to the highly increased computational cost of attention mechanism during long-term temporal modeling,
DTA is only applied to a short-term segment of the whole video.
To achieve long-range temporal alignment, 
we propose to adaptively store and compress the historical segment information in a cache unit
and compensate the DTA module with the cached information.

Specifically, we first initialize the cache unit with a zero tensor $F_{cache} \in \mathbb{R}^{l \times h \times w \times d}$,
where $l$ is the cache size, and $h$, $w$ are the height and width of the feature map.
We define the AMC module with two key operations: query and update.

For cache querying, 
we use a cross-attention module,
which shares parameters with temporal attention in the DTA module,
to retrieve information from the cache unit $F_{cache}$..
Then we fuse the queried tokens with the output feature $F_{out}^i$ of DTA.
the output of cache attention is fused with the output feature $F_{out}$ of DTA.

For cache updating, 
the input feature from the new segment $F \in \mathbb{R}^{n \times h \times w \times d}$ 
is first downsampled via average pooling to obtain $F_{pool} \in \mathbb{R}^{l \times h \times w \times d}$.  
Then, we design a gating mechanism to control the update of the cache unit $F_{cache}$, which is implemented as follows:
\begin{align}
    g &= \sigma(W_{gate}([F_{pool}, F_{cache}])) \\
    F_{cache} &= (1 - g) \odot F_{cache} + g \odot F_{pool}
\end{align}

where $\sigma$ denotes the sigmoid function,
$[\cdot, \cdot]$ denotes concatenation operation and
$W_{gate} \in \mathbb{R}^{2d \times d}$ is a linear projection matrix.
The gating mechanism adaptively controls how much of the new information is incorporated to update the cache unit.

\begin{figure}[h] %
    \centering %
    \includegraphics[width=0.75\textwidth]{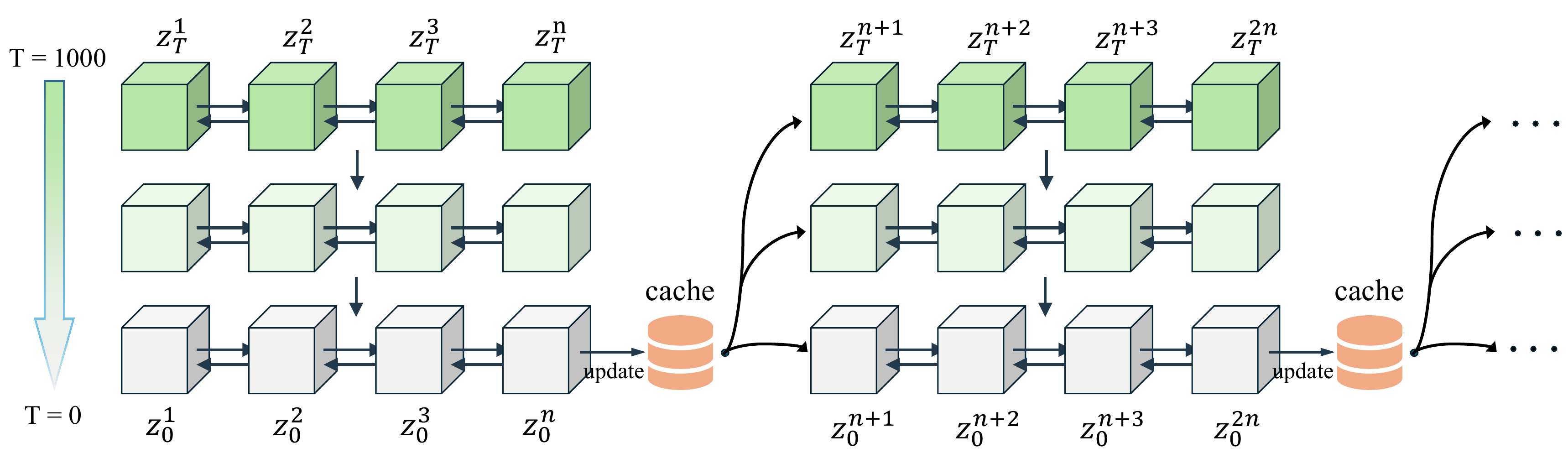}
    \caption{
        An illustration of the inference process.
        The input video is divided into multiple segments, where $n$ indicates segment length. 
        The cache from the last denoising step is then used to guide the generation of the subsequent segment.
    } %
    \label{fig:sampling} %
\end{figure}

\subsection{Asymmetric Sampling Strategy}
Inspired by \cite{df}, we introduce an asymmetric sampling strategy 
that mitigates feature mismatches arising from different diffusion sampling steps to 
stablize flexible cache interaction across segments.
Specifically, this strategy diffuses different segments with distinct timesteps
during training as shown in Fig. \ref{fig:model}, allowing model to adapt to cache interactions at varying noise levels.

We provide a detailed illustration of the inference process in Fig. \ref{fig:sampling}.
The input video is divided into multiple segments for sequential processing 
and we leverage the attention memory cache from the last denoising step
to guide the generation of the subsequent segment.

This strategy not only reduces the memory footprint of the AMC module,
which would otherwise store features from all denoising steps of the current segment,
but also provides more informative guidance and avoids noise accumulation, 
as features from later denoising steps are less noisy and contain richer detail.

\section{Experiments}
\label{sec:exp}
\subsection{Experimental Settings}
\paragraph{Training and Testing Datasets.}
We train LiftVSR on both image and video datasets covering a wide range of natural scenes and facial content.
The image datasets include DF2K~\cite{div2k,flicker2k}, LSDIR \cite{lsdir} and the first 10K face images from FFHQ \cite{ffhq}.
The video training set is composed of the following datasets: 
1) REDS \cite{reds} dataset. 
2) YouHQ \cite{uav} dataset, consisting of 37K high-quality (1080×1920) video clips.
3) A subset of OpenVid-1M~\cite{openvid} dataset, containing \textasciitilde120K video-text pairs with a minimum resolution of 512×512.
We follow the degradation pipeline proposed in RealBasicVSR to synthesize LQ-HQ training pairs.
For evaluation, we adopt four commonly used synthetic benchmarks (REDS30, SPMCS~\cite{spmcs}, UDM10~\cite{udm}, YouHQ40) 
and the real-world benchmarks VideoLQ~\cite{realbasicvsr} to assess model performance.

\begin{table}[h]
  \centering
  \caption{
    Quantitative results of our method and the state-of-the-art methods on both
    synthetic (UDM10, REDS30, YouHQ40, SPMCS) and real-world (VideoLQ) benchmark. 
    The best and second results are highlighted in \textbf{bold} and \underline{underline}, respectively.
    $E^*_{warp}$ denotes $E_{warp}$ ($\times10^{-3}$).
  }
  \label{tab:main_table}
  \resizebox{\textwidth}{!}{
  \renewcommand{\arraystretch}{1.1}
  \begin{tabular}{cc|ccc|cccccc}
  \toprule
  Datasets & Metrics & Real-ESRGAN & RealBasicVSR & RealViformer & StableSR & STAR & UAV & MGLD-VSR & Ours \\ 
  \midrule
  \midrule
  \multirow{5}{*}{UDM10} & PSNR$\uparrow$ & 27.899 & \underline{28.548} & \textbf{29.557} & 26.380 & 27.255 & 28.111 & 28.486 & 27.654 \\
  ~ & MUSIQ$\uparrow$ & 59.463 & \underline{62.590} & 59.398 & 56.634 & 41.245 & 33.415 & 62.339 & \textbf{65.988}  \\
  ~ & CLIPIQA$\uparrow$ & 0.4909 & 0.4402 & 0.4039 & \underline{0.5346} & 0.2743 & 0.2190 & 0.4774 & \textbf{0.5358} \\
  ~ & DOVER$\uparrow$ & \underline{0.4840} & 0.4800 & 0.4710 & 0.4682 & 0.4501 & 0.3481 & 0.4728 & \textbf{0.5309} \\
  ~ & $E^*_{warp}$ $\downarrow$ & 0.229 & 0.125 & 0.105 & 0.600 & \textbf{0.078} & \underline{0.115} & 0.228 & 0.197 \\ 
  \midrule
  \multirow{5}{*}{REDS30} & PSNR$\uparrow$ & 23.910 & 25.209 & \textbf{26.005} & 23.547 & 24.351 & 25.110 & \underline{25.210} & 24.345  \\
  ~ & MUSIQ$\uparrow$ & \underline{66.359} & 65.100 & 62.166 & 61.073 & 38.117 & 23.276 & 64.463 & \textbf{68.052} \\
  ~ & CLIPIQA$\uparrow$ & \underline{0.4742} & 0.3732 & 0.3425 & 0.4348 & 0.2290 & 0.1556 & 0.3935 & \textbf{0.4851} \\
  ~ & DOVER$\uparrow$ & 0.38076 & 0.3622 & 0.3554 & 0.3537 & 0.3727 & 0.2295 & \underline{0.3872} & \textbf{0.4071}  \\
  ~ & $E^*_{warp}$ $\downarrow$ & 0.806 & 0.396 & 0.302 & 1.022 & \underline{0.275} & \textbf{0.179} & 0.608 & 0.613 \\ 
  \midrule
  \multirow{5}{*}{YouHQ40} & PSNR$\uparrow$ & 25.717 & 25.217 & \textbf{26.770} & 23.298 & 26.150 & \underline{26.229} & 26.081 & 25.590 \\
  ~ & MUSIQ$\uparrow$ & 61.186 & \underline{66.463} & 64.217 & 62.994  & 41.695 & 31.258 & 63.378 & \textbf{66.638} \\
  ~ & CLIPIQA$\uparrow$ & 0.4883 & 0.4872 & 0.4578 & \underline{0.5145} & 0.3179 & 0.2243 & 0.4699 & \textbf{0.5611} \\
  ~ & DOVER$\uparrow$ & \underline{0.6106} & 0.5882 & 0.5837 & 0.5765 & 0.5929 & 0.4583 & 0.5858 & \textbf{0.6156}  \\
  ~ & $E^*_{warp}$ $\downarrow$ & 0.466 & 0.276 & 0.214 & 0.847 & \textbf{0.093} & \underline{0.186} & 0.286 & 0.241 \\ 
  \midrule
  \multirow{5}{*}{SPMCS} 
    & PSNR$\uparrow$    & 23.686           & \underline{24.623}  & \textbf{25.257}   & 22.527  & 22.321  & 23.687 & 23.470  & 24.041 \\
  ~ & MUSIQ$\uparrow$   & 66.284          & \underline{67.698} & 64.921           & 61.687 & 38.711 & 34.866 & 65.728 & \textbf{70.144} \\
  ~ & CLIPIQA$\uparrow$ & 0.5256          & 0.4123 & 0.3878           & 0.5211 & 0.2685 & 0.2130 & 0.4333 & \textbf{0.5407} \\
  ~ & DOVER$\uparrow$   & 0.4411 & 0.4530 & 0.4417 & 0.4447 & 0.3666 & 0.3359 & \underline{0.4543} & \textbf{0.5168} \\
  ~ & $E^*_{warp}$ $\downarrow$ & 0.754 & 0.174 & 0.146 & 0.688 & \underline{0.094} & \textbf{0.094} & 0.162 & 0.132 \\ 
  \midrule
  \midrule
  \multirow{3}{*}{VideoLQ}
    & MUSIQ$\uparrow$   & 49.849 & \underline{55.594} & 52.135 & 48.525 & 41.1390 & 38.439 & 51.197 & \textbf{56.455} \\
  ~ & CLIPIQA$\uparrow$ & 0.3519 & \textbf{0.3879}  & 0.3464 & 0.4057 & 0.2848  & 0.2482 & 0.3476 & \underline{0.3565} \\
  ~ & DOVER$\uparrow$   & 0.4383 & 0.45839 & 0.4252 & 0.3984 & 0.4633  & 0.3729 & \underline{0.4640} & \textbf{0.4819} \\
  \bottomrule 
  \end{tabular}
  }
\end{table}

\paragraph{Implementation Details.}
We use PixArt-$\alpha$~\cite{pixart-alpha} as the backbone for our model, which is a lightweight and efficient architecture.
The training process is divided into two stages: 
1) We first pre-train the model on image datasets with a batch size of 64.
2) We then insert the designed temporal modules to model and train them on the video datasets with a batch size of 8.
Both stages are trained on 4$\times$RTX 4090 GPUs with a learning rate of 1e-5.
We set the segment length processed by the DTA module to 8 frames, 
and the total video length during training is set to a multiple of 8.

\paragraph{Evaluation Metrics.}
Following the common practice, 
we use various metrics to evaluate both the frame quality and temporal consistency.
For synthetic datasets with LQ-HQ pairs, we employ PSNR, and the flow warping error $E_{warp}$ \cite{liu2024evalcrafter}
and no-reference perceptual metrics including MUSIQ \cite{musiq}, CLIPIQA \cite{clipiqa} and DOVER \cite{dover} for evaluation.
Notably,
despite the widespread use of $E_{warp}$ in various papers \cite{uav,diffvsr,star}, it tends to produce lower values for
blurry and smooth video sequences \cite{mgldvsr}. Therefore, relying solely on $E_{warp}$ to assess temporal consistency is not reasonable. Thus we also provide a temporal profile comparison in Fig. \ref{fig:temporal_profile} to truly reflect the model performance.
For real-world video data, we use three no-reference metrics: MUSIQ, CLIPIQA and DOVER.

\subsection{Comparisons}
To demonstrate the effectiveness of our method, 
we compare it with several state-of-the-art methods, including
Real-ESRGAN \cite{realesrgan}, RealBasicVSR \cite{realbasicvsr}, 
RealViformer \cite{realviformer}, StableSR \cite{stablesr}, 
STAR~\cite{star}, Upscale-A-Video (UAV) \cite{uav} and MGLD-VSR \cite{mgldvsr}.

\paragraph{Quantitative Comparison.}
We present the quantitative results on both synthetic and real-world datasets in Table~\ref{tab:main_table}.
On the four synthetic test sets, 
our method significantly outperforms all existing approaches in perceptual image quality metrics like MUSIQ and CLIPIQA,
as well as in the video quality metric DOVER. 
We also achieve competitive results in PSNR metrics, although slightly lower than non-generative methods like RealViformer \cite{realviformer}.
These results demonstrate that LiftVSR is capable of generating high-quality videos with rich details and textures
while maintaining high fidelity to the original content.
On the real-world dataset VideoLQ \cite{realbasicvsr}, 
we only using three perceptual metrics (MUSIQ, CLIPIQA, DOVER) for evaluation.
We also achieve the best performance in MUSIQ and DOVER,
and second best in CLIPIQA, 
further validating the robustness and effectiveness of LiftVSR in handling complex real-world degradations.

\begin{figure}[h] %
    \centering %
    \includegraphics[width=\textwidth]{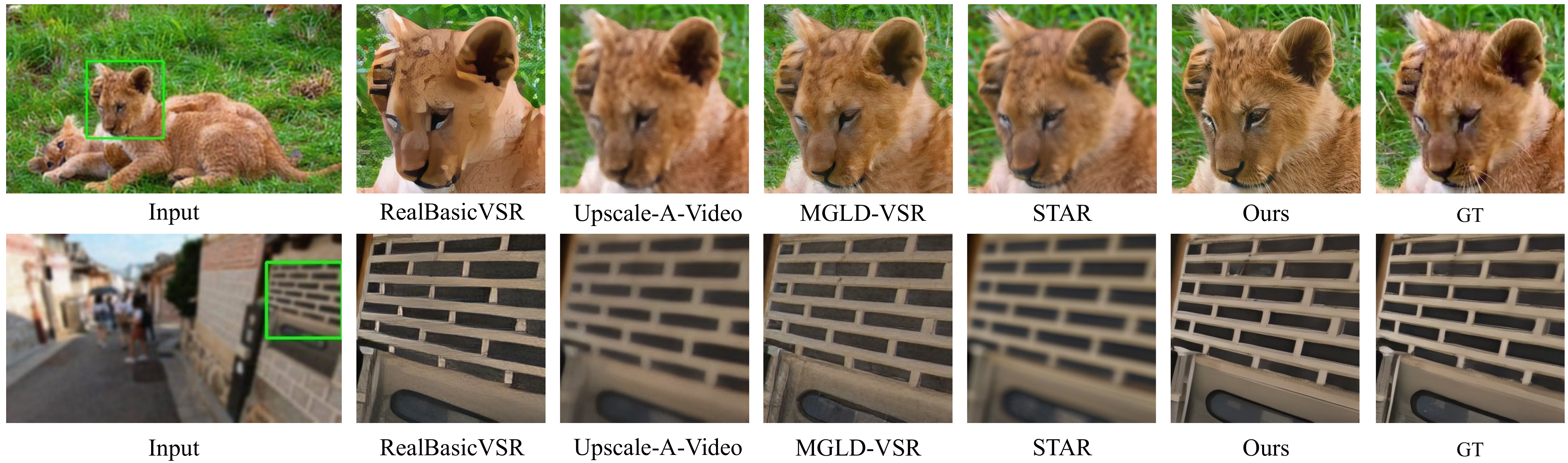}
    \caption{
        Visual comparison of our method with state-of-the-art methods 
        on synthetic low-quality videos from YouHQ40 and REDS30 datasets.
        \textbf{(Zoom-in for best view)}
    } %
    \label{fig:comparison1} %
\end{figure}

\begin{figure}[h] %
    \centering %
    \includegraphics[width=\textwidth]{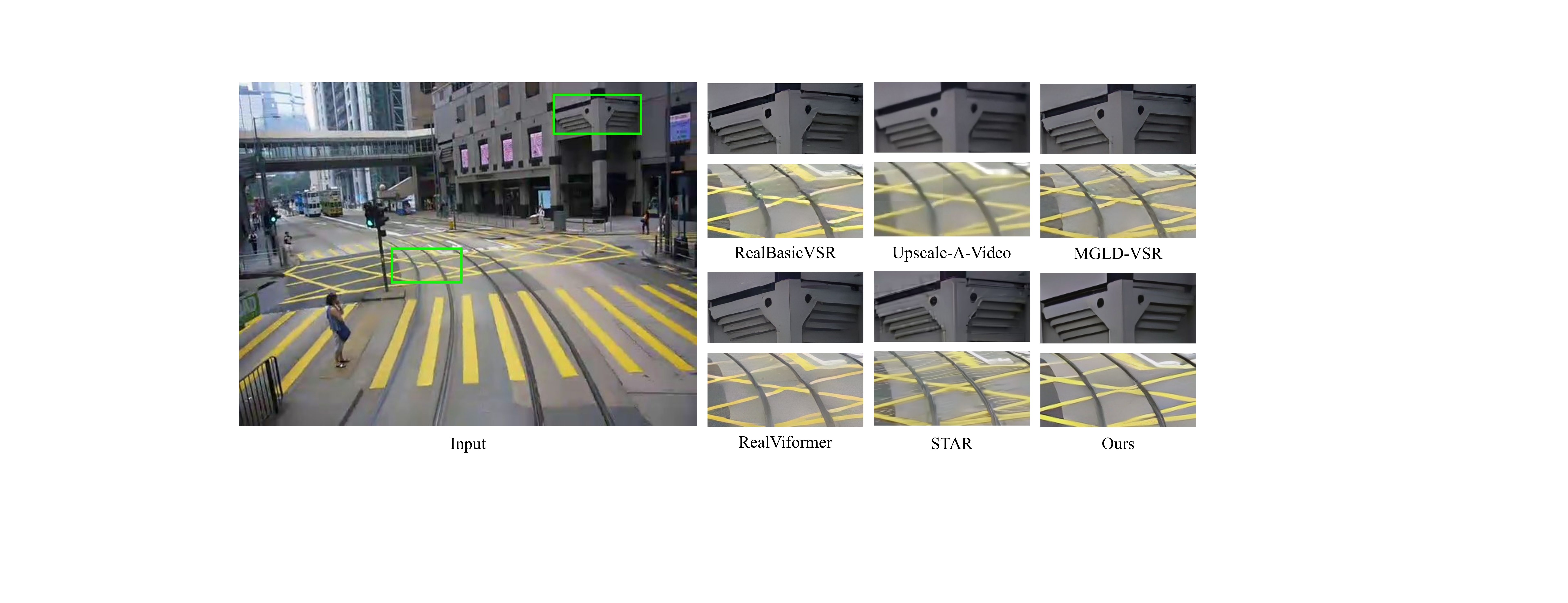}
    \caption{
        Visual comparison of our method with state-of-the-art methods 
        on real-world videos from VideoLQ \cite{realbasicvsr} datasets.
        LiftVSR effectively removes mixed distortions from the input,
        generating clean and high-quality videos than other SOTA methods.
        \textbf{(Zoom-in for best view)}
    } %
    \label{fig:comparison2} %
\end{figure}

\paragraph{Qualitative Comparison.}
We present visual comparisons on both synthetic and real-world datasets 
in Fig. \ref{fig:comparison1} and Fig. \ref{fig:comparison2}, respectively. 
The results in Fig. \ref{fig:comparison1} are selected from the YouHQ40~\cite{uav} and REDS30~\cite{reds} datasets,
showcases our method's excellence in detail reconstruction, particularly in texture and edge restoration.
Specifically, in the first row, the details of animal fur generated by LiftVSR 
are more vivid and clear, with a more natural and smooth background. 
In contrast, other methods tend to produce blurry and unnatural textures. 
In the second row, we also achieve higher fidelity to the original content than the others, 
accurately restoring the texture of the window edge.
For real-world scenarios, LiftVSR achieve superior degradation handling capabilities. 
As shown in Fig. \ref{fig:comparison2}, LiftVSR effectively removes mixed distortions from the input, 
restoring true details. Other methods still face challenges in dealing with complex distortions in real-world scenarios.
The both quantitative and qualitative results demonstrate the effectiveness of our method in generating high-quality videos
and the robustness in handling complex real-world degradations.
More experimental comparisons are provided in the supplementary materials.

\begin{figure}[h] %
    \centering %
    \includegraphics[width=\textwidth]{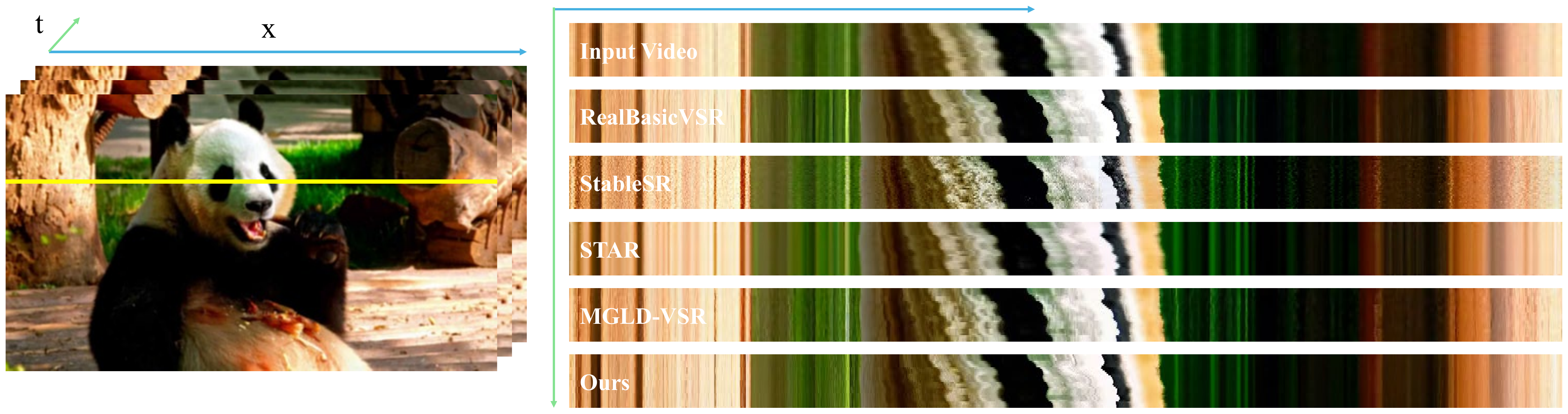}
    \caption{
        Temporal profiles comparision.
        We select a row of pixels from the generated video and
        stack the pixel values over time.
        As shown in figure, the profile of single-frame models like StableSR \cite{stablesr}
        exhibit poor temporal stability. Compared to other methods,
        LiftVSR shows excellent continuity in the temporal dimension as well as rich details and textures.
    } 
    \label{fig:temporal_profile} 
\end{figure}

\paragraph{Temporal Consistency.}
Benefiting from the hybrid temporal modeling, 
we achieve a significant improvement in temporal consistency.
Although $E_{warp}$ does not accurately reflect the temporal consistency of the results \cite{mgldvsr}, we still show it in Tab. \ref{tab:main_table} as a reference.
Our method tends to generate more details and textures, resulting in higher $E_{warp}$ compared to methods like UAV~\cite{uav} and STAR~\cite{star}
as shown in Tab. \ref{tab:main_table}.
To truly illustrate the temporal consistency of our method, 
we present a subjective comparison through a temporal profile comparison.
As shown in Fig. \ref{fig:temporal_profile}, single-frame models like StableSR \cite{stablesr} exhibit poor temporal stability. 
In contrast, LiftVSR shows excellent continuity in the temporal dimension
compared to other methods,
achieving high temporal coherence along with richer details and textures.

\subsection{Ablation Study}
To better demonstrate the effectiveness of our hybrid temporal modeling mechanism, 
we conducted ablation studies under different settings. 
We compared the impact of the DTA, AMC module and the asymmetric sampling strategy (ASS) on the performance of our method as shown in Tab.~\ref{tab:ablation}.
More ablation experiments and visual comparisons are provided in the supplement materials.

\begin{table}[h]
    \centering
    \caption{
        Ablation study of different components in our method, test on YouHQ40 dataset.
    }
    \label{tab:ablation}
    \resizebox{0.65\textwidth}{!}{
    \renewcommand{\arraystretch}{0.9}
    \begin{tabular}{c|ccc|cccc}
    \toprule
    Exp. & DTA & AMC & ASS & PSNR$\uparrow$ & MUSIQ$\uparrow$ & DOVER$\uparrow$ & $E^*_{warp}$$\downarrow$ \\ 
    \midrule
    (a) &              &              &              & 24.921 & 67.256 & 0.5675 & 0.325 \\
    (b) & $\checkmark$ &              &              & 25.309 & \textbf{67.328} & 0.5766 & 0.295 \\
    (c) & $\checkmark$ & $\checkmark$ &              & 25.541 & 66.282& 0.5971 & 0.258 \\
    (d) & $\checkmark$ & $\checkmark$ & $\checkmark$ & \textbf{25.590} & 66.638 & \textbf{0.6156} & \textbf{0.241} \\
    \bottomrule 
    \end{tabular}
    }
\end{table}

\paragraph{Effectiveness of Dynamic Temporal Attention.}
The introduced DTA module is designed to enhance the fine-grained temporal modeling capability of our method. 
As shown in Table \ref{tab:ablation}, Exp. (a) and (b) illustrate the effects of using conventional temporal attention and the DTA module, respectively. 
The DOVER score and $E_{warp}$ metric in Exp. (b) is significantly improved compared to Exp. (a),
demonstrating effective enhancement in temporal modeling.

\paragraph{Effectiveness of Attention Memory Cache.}
In addition to DTA module, we also propose AMC module to maintain long-term temporal coherence.
We can observe that the DOVER score and $E_{warp}$ metric in Exp. (c) obtain further improvement.
It also helps improve PSNR score, due to the better temporal coherence and less noise in the generated video.

\paragraph{Effectiveness of Asymmetric Sampling Strategy.}
The asymmetric sampling strategy aims to further 
stablize cache interaction during inference.
Exp. (d) obtain the best performance over all experiments, with all proposed modules enabled.
The increased DOVER score and reduced $E_{warp}$ metric in Exp. (d) demonstrate the effectiveness of the ASS module.

\section{Conclusion}
\label{sec:conclusion}
In this paper, we present LiftVSR, a novel framework aims to
lift image diffusion model to video super-resolution task
with lower computational cost and higher video quality.
We propose a hybrid temporal modeling mechanism, including two key components:
1) Dynamic Temporal Attention for robust and efficient short-term temporal modeling,
2) Attention Memory Cache for long-term temporal feature propagation.
The introduced asymmetric sampling strategy is further designed to 
stablize the cache interaction during inference by mitigating feature mismatches arising from
different sampling steps.
Extensive experiments on multiple synthetic and real-world benchmarks demonstrate our
superior performance over other state-of-the-art methods and the effectiveness of our proposed components.
The efficient design of LiftVSR will also benefit the practical applications and future research.

\bibliographystyle{plain}
\bibliography{ref}

\newpage
\appendix

\begin{center}
  \Large{\textbf{Appendix}}
\end{center}
\section{Architecture}
\subsection{More detials on Training}
Our proposed LiftVSR is constructed based on the Text-to-Image (T2I) model PixArt-$\alpha$~\cite{pixart-alpha}, 
which is a latent diffusion model (LDM)~\cite{ldm} built upon the DiT~\cite{dit} architecture. 
The architecture of LiftVSR comprises four main components: 
text encoder, VAE, ControlNet and DiT model with Dynamic Temporal Attention (DTA) and Attention Memory Cache (AMC) modules.
The architecture of the ControlNet model is inspired by~\cite{pixart-delta}, 
and it is used to incorporate low-quality video conditions.
In Table~\ref{tab:params}, we provide a detailed information of the 
pretrained DiT model and inserted temporal modules.

\begin{table}[h]
  \centering
  \caption{
    Hyperparameters for our proposed LiftVSR.
  }
  \label{tab:params}
  \resizebox{0.6\textwidth}{!}{
  \begin{tabular}{ccc}
  \toprule
  Module & Hyperparameter & Value \\
  \midrule
  \multirow{6}{*}{DiT} 
  & Training patch shape & 8$\times$3$\times$512$\times$512 \\
  & z-shape& 4$\times$64$\times$64 \\
  & Patch size & 2 \\
  & Channels & 1152 \\
  & Head number & 16 \\
  & Block number & 28 \\
  \midrule

  \multirow{1}{*}{ControlNet}
  & Block number & 12 \\
  \midrule

  \multirow{3}{*}{VAE}
  & $f$ & 8 \\
  & Channels & 128 \\
  & Channel multiplier & 1, 2, 4, 4 \\
  \midrule

  \multirow{3}{*}{DTA} 
  & Block interval & 3 \\
  & Head number & 16 \\
  & Positional encoding & RoPE~\cite{rope} \\
  \midrule

  \multirow{1}{*}{AMC} 
  & Memory cache length & 2 \\
  \bottomrule
  \end{tabular}
  }
\end{table}

\subsection{More detials on Inference}
\paragraph{Video VAE.}
We found that using the original VAE from PixArt-$\alpha$~\cite{pixart-alpha} 
to decode the input video latents results in significant temporal flickering and artifacts in the output.
This is attributed to the original VAE is trained solely on image data. 
To mitigate this issue, we introduced a Video VAE architecture to further enhance low-level consistency. 
Specifically, we investigated the Video VAE model utilized in Stable Video Diffusion (SVD)~\cite{svd}, 
which has been trained on a large amount of video data. 
Given that the Video VAE in SVD shares the same encoder architecture and parameters with the VAE in PixArt-$\alpha$,
we were able to seamlessly integrate the Video VAE decoder from SVD into our LiftVSR framework.
The Video VAE effectively mitigates the temporal flickering and reduce to warping error as shown in Table~\ref{tab:vae}.

\begin{table}[b]
  \centering
  \caption{
      Ablation study of different components during inference (on YouHQ40 dataset).
  }
  \label{tab:vae}
  \resizebox{\textwidth}{!}{
  \renewcommand{\arraystretch}{0.9}
  \begin{tabular}{c|cc|ccccc}
  \toprule
  Exp. & Video VAE & Color Correction & PSNR$\uparrow$ & MUSIQ$\uparrow$ & CLIPIQA$\uparrow$ & DOVER$\uparrow$ & $E^*_{warp}$$\downarrow$ \\ 
  \midrule
  (a) &  $\checkmark$ & & 25.028 & 66.721 & 0.5592 & 0.6201 & 0.246 \\
  (b) &  & $\checkmark$ & 24.917 & 67.747 & 0.5763 & 0.6134 & 0.475\\
  (v) & $\checkmark$ & $\checkmark$ & 25.590 &  66.638 &  0.5611 &  0.6156 & 0.241 \\

  \bottomrule 
  \end{tabular}
  }
\end{table}

\paragraph{Inference at Arbitrary Resolution.}
During inference, to accommodate inputs of arbitrary resolution, 
we partition the input video into multiple overlapping patches of size 512$\times$512. 
We then employ the sampling strategy proposed by \cite{stablesr} to seamlessly merge these overlapping patches. 
For inputs of arbitrary length, we divide the input video into multiple temporal segments. 
The proposed Attention Memory Cache (AMC) module will maintain temporal coherence across these segments. 
Additionally, we apply an overlapping strategy in the temporal dimension, 
which further enhances temporal coherence. We set the overlap length to one frame.

\paragraph{Color Correction.}
Following previous works~\cite{stablesr,uav},
we apply color correction to avoid color shift artifacts in the generated video.
Specifically, we perform color normalization~\cite{stablesr} on the generated video to align its mean and variance with those of the LQ input.
We can observe that color correction can effectively improve the PSNR scores as shown in Table~\ref{tab:vae}.

\begin{figure}[t] %
  \centering %
  \includegraphics[width=\textwidth]{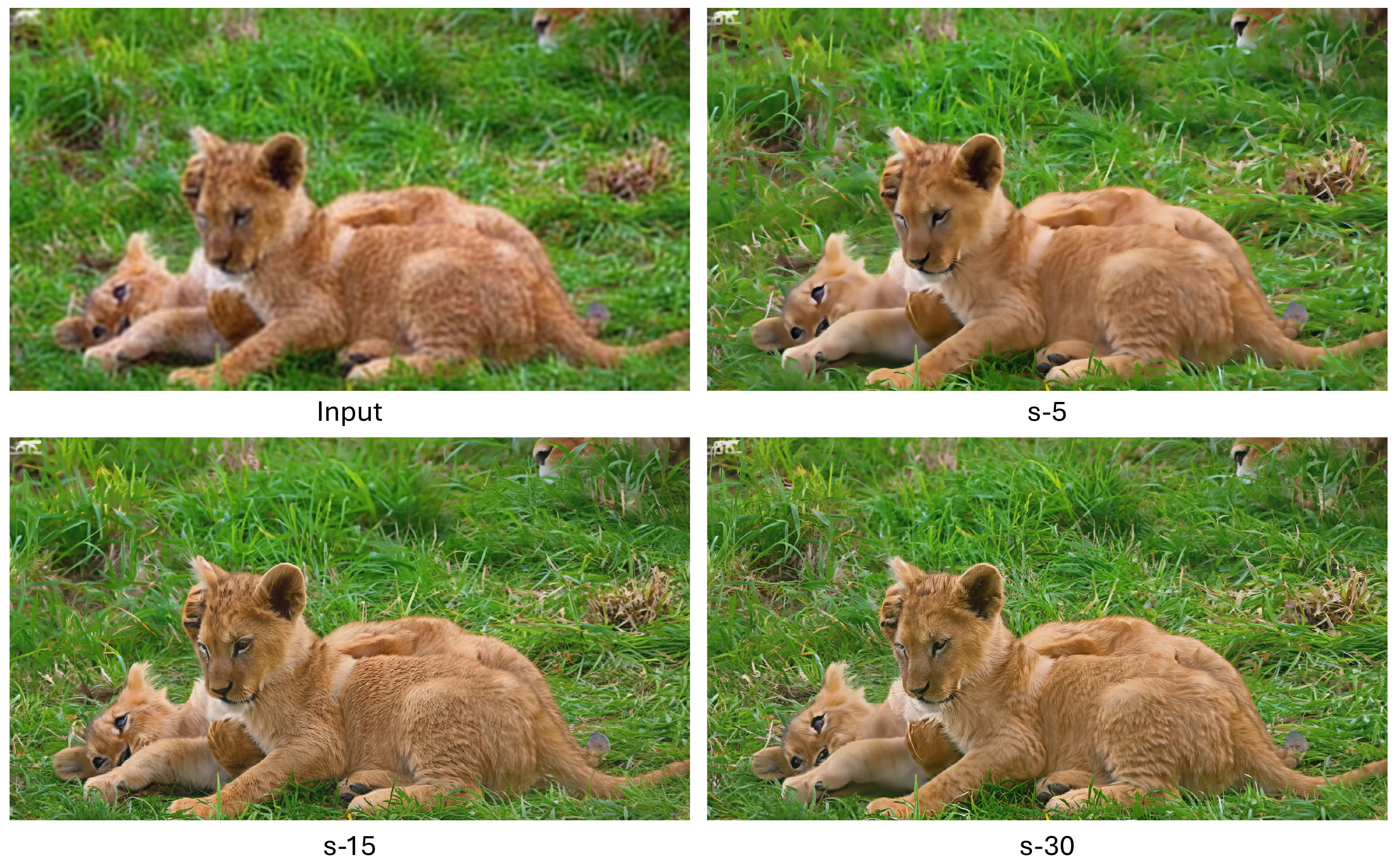}
  \caption{Comparison of results with different sampling steps.\textbf{(Zoom-in for best view)}}
  \label{fig:steps} 
\end{figure}

\paragraph{Sampling Steps.}
Diffusion models define a multi-step process of noise addition and removal. 
In this study, the total number of noise addition steps is set to $ T = 1000 $. 
During inference, various accelerated sampling methods \cite{ddim,lu2022dpm} can be employed, 
In our experiments, we default to using DPM-Solver~\cite{lu2022dpm} for sampling,
and we observed that a higher sampling steps generates more detailed texture, 
while a lower steps leads to blurrier and smoother outcomes. 
To balance speed and quality, we set the sampling steps to 15 in all our experiments,
which efficiently produced high-quality results. 
The comparison of results with different sampling steps is illustrated in Figure~\ref{fig:steps}.

\section{More Results}
\subsection{More Quantitative Comparisons}
In this section, we present additional quantitative comparisons on both synthetic and real-world benchmarks 
against state-of-the-art methods, as detailed in Table~\ref{tab:main_table}. 
We achieve competitive performance on the non-reference metric IL-NIQE~\cite{il-niqe}, demonstrate the 
strong perceptual quality of our results.
For the reference metrics SSIM and LPIPS~\cite{lpips}, our results are slightly lower than those of non-generative methods 
such as RealViformer~\cite{realviformer}, which can be attributed to the inherent stochastic nature of diffusion models.

\begin{table}[t]
  \centering
  \caption{
    Additional quantitative results of our method and the state-of-the-art methods.
    The best and second results are highlighted in \textbf{bold} and \underline{underline}, respectively.
  }
  \label{tab:main_table}
  \resizebox{\textwidth}{!}{
  \renewcommand{\arraystretch}{1.1}
  \begin{tabular}{cc|ccc|cccccc}
  \toprule
  Datasets & Metrics & Real-ESRGAN & RealBasicVSR & RealViformer & StableSR & STAR & UAV & MGLD-VSR & Ours \\ 
  \midrule
  \midrule
  \multirow{3}{*}{UDM10}
  ~ & SSIM$\uparrow$ & \underline{0.8286} & 0.8254 & \textbf{0.8515} & 0.7431 & 0.8100 & 0.8122 & 0.8192 & 0.7885  \\
  ~ & LPIPS$\downarrow$ & 0.2701 & 0.2666 & \textbf{0.2338} & 0.3263 & 0.3027 & 0.3410 & \underline{0.2428} & 0.2923  \\
  ~ & IL-NIQE$\downarrow$ & 30.696 & 28.697 & 27.944 & 26.150 & 35.843 & 36.172 & \textbf{25.364} & \underline{26.002} \\ 
  \midrule
  \multirow{3}{*}{REDS30}
  ~ & SSIM$\uparrow$ & 0.6287 & \underline{0.6546} & \textbf{0.6835} & 0.5853 & 0.6324 & 0.6424 & 0.6466 & 0.6086 \\
  ~ & LPIPS$\downarrow$ & 0.3088 & 0.2543 & \textbf{0.2233} & 0.3114 & 0.4557 & 0.4794 & \underline{0.2041} & 0.3468 \\
  ~ & IL-NIQE$\downarrow$ & \underline{19.668} & 19.801 & 19.964 & 19.826 & 25.296 & 26.250 & \textbf{18.567} & 20.399\\ 
  \midrule
  \multirow{3}{*}{YouHQ40}
  ~ & SSIM$\uparrow$ & 0.7067 & 0.6778 & \textbf{0.7208} & 0.6338  & \underline{0.7146} & 0.7047 & 0.6899 & 0.6757 \\
  ~ & LPIPS$\downarrow$ & 0.3200 & 0.3466 & \textbf{0.2896} &  0.5275 & 0.3631 & 0.3958 & \underline{0.3120} & 0.3253 \\
  ~ & IL-NIQE$\downarrow$ & 24.885 & \underline{22.877} & 23.439 & 23.034 & 30.109 & 32.421 & 23.498 & \textbf{22.521}\\ 
  \midrule
  \multirow{3}{*}{SPMCS} 
  ~ & SSIM$\uparrow$    & 0.6201          & 0.6264 & \textbf{0.6632}  & 0.5913 & 0.6009 & 0.6352 & \underline{0.6372} & 0.6111  \\
  ~ & LPIPS$\downarrow$ & 0.3496          & \underline{0.3317} & \textbf{0.2959}  & 0.5191 & 0.6377 & 0.5889 & 0.4723 & 0.3557 \\
  ~ & IL-NIQE$\downarrow$ & 25.403 & 25.890 & 26.332 & 25.644 & 34.206 & 37.157 & \underline{25.032} & \textbf{24.201} \\ 
  \midrule
  \midrule
  \multirow{1}{*}{VideoLQ}
  ~ & IL-NIQE$\downarrow$ & 27.943 & 26.290 & 26.225 & 25.909 & 29.678 & 31.118 & \textbf{24.059} & \underline{26.219}\\
  \bottomrule 
  \end{tabular}
  }
\end{table}

\subsection{Effectiveness of Attention Memory Cache}
\begin{figure}[h] %
  \centering %
  \includegraphics[width=\textwidth]{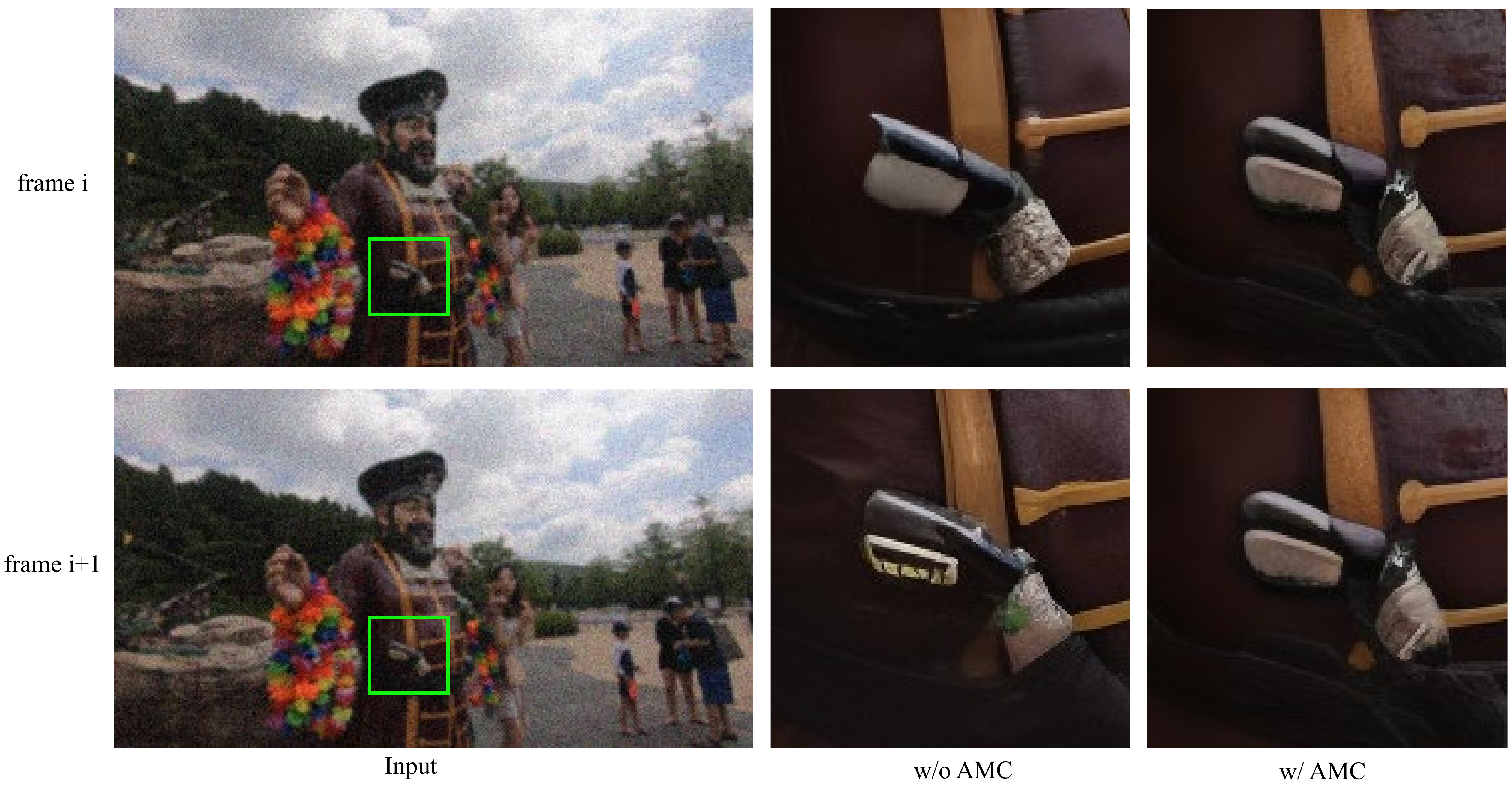}
  \caption{
    Comparison of results across different segments with and without AMC module.}
  \label{fig:amc} 
\end{figure}

The Attention Memory Cache (AMC) module is specifically designed to maintain the temporal coherence across different video segments. 
In the main paper, we conducted a quantitative ablation study to evaluate the effectiveness of the AMC module. 
To further illustrate its impact, we provide qualitative comparisons in Figure~\ref{fig:amc}.
The generated results across different segments demonstrate that the AMC module effectively maintains low-level consistency across segments.

\subsection{More Qualitative Comparisons}
\begin{figure}[h] %
  \centering %
  \includegraphics[width=\textwidth]{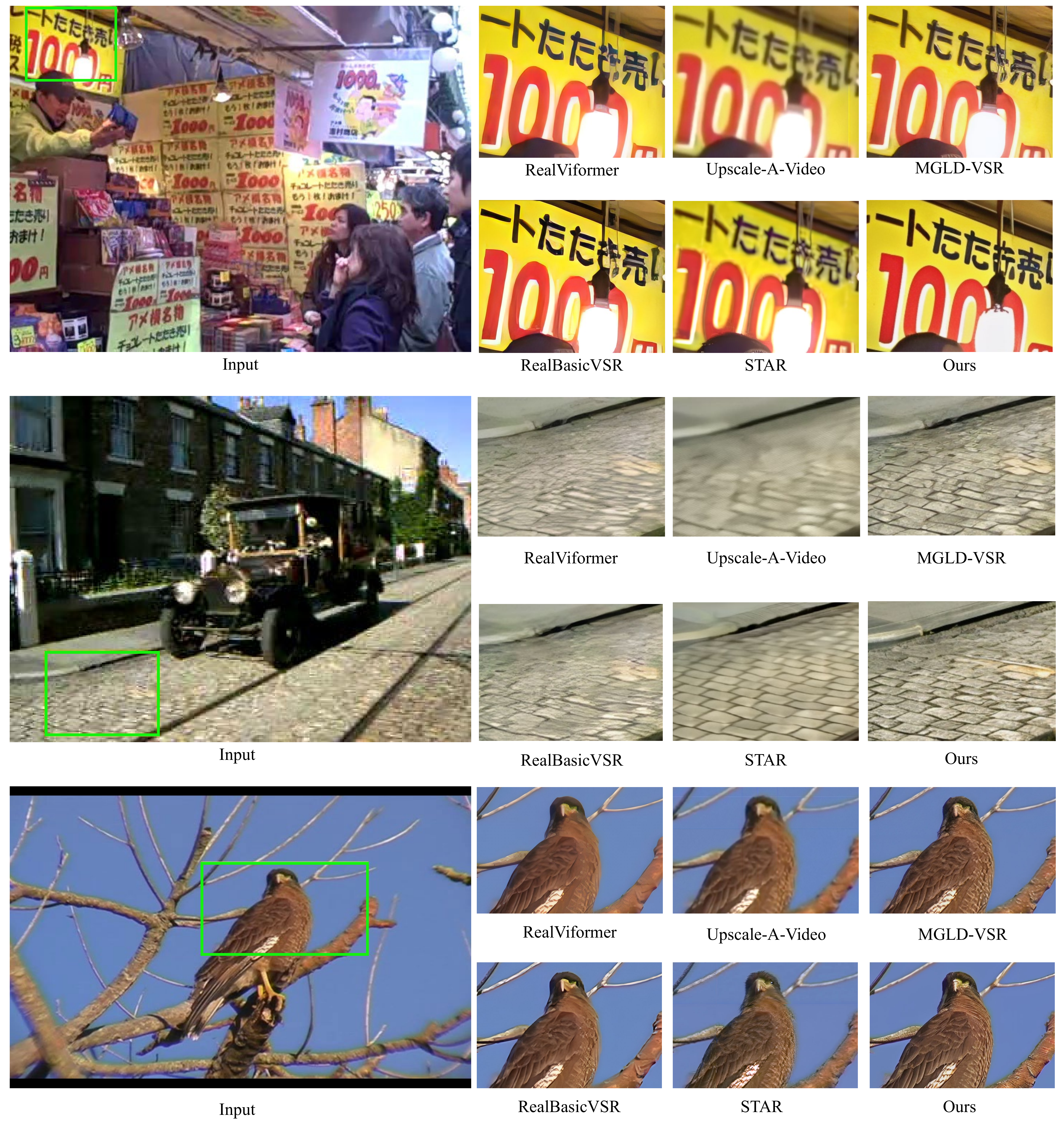}
  \caption{Qualitative comparisons on real-world videos.\textbf{(Zoom-in for best view)}}
  \label{fig:more_compare1} 
\end{figure}

\begin{figure}[h] %
  \centering %
  \includegraphics[width=\textwidth]{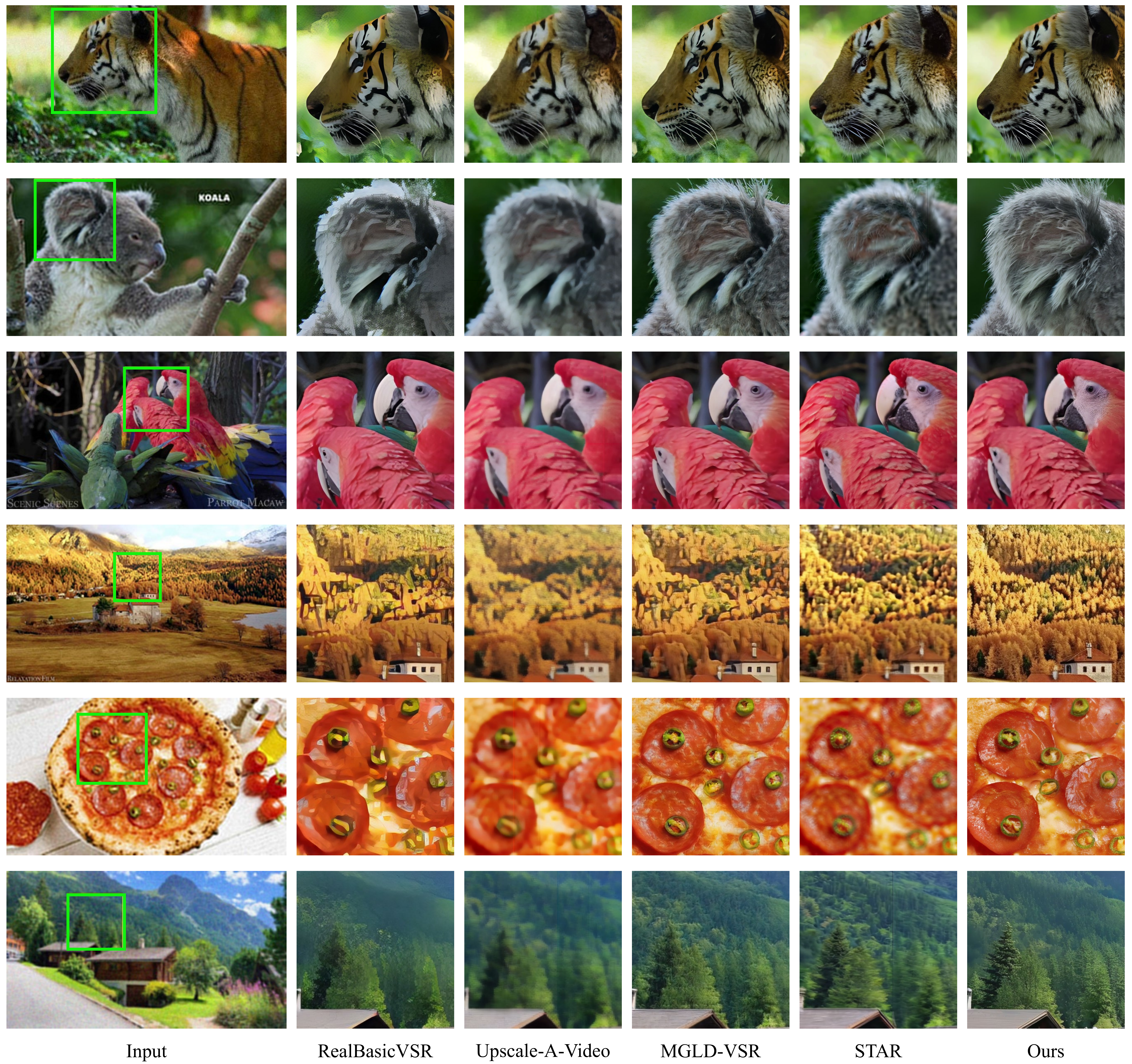}
  \caption{Qualitative comparisons on synthetic low-quality videos.\textbf{(Zoom-in for best view)}}
  \label{fig:more_compare2} 
\end{figure}

In this section, we provide more qualitative comparisons on synthetic and real-world benchmarks
as shown in Figure~\ref{fig:more_compare1} and Figure~\ref{fig:more_compare2}.

\subsection{Video Demo}
\label{sec:demo}
We also provide several generated videos in the supplemental materials to demonstrate the effectiveness of our proposed LiftVSR. However, due to the size constraints of the supplemental materials, the videos have been compressed and may not fully reflect the optimal results.

\section{Limitations}
Despite the significant performance improvements that LiftVSR has achieved in both synthetic 
and real-world video super-resolution tasks, several limitations remain. 
First, for small objects, faces, and text scenes in the input video, 
LiftVSR may not be able to accurately reconstruct them due to the inherent stochastic nature of diffusion models.
Second, when processing high-resolution and diverse aspect-ratio inputs, 
the model requires an overlapping sampling strategy, which reduces inference efficiency. 
Future work could explore enabling the model to directly support arbitrary-resolution video inputs, 
thereby further enhancing efficiency. 
Third, although the inference speed of LiftVSR is much faster than other diffusion-based methods~\cite{uav,star,mgldvsr}, 
it still falls short of real-time performance. 
Future efforts may focus on techniques such as model pruning and distillation to reduce the number of parameters and sampling steps,
further improving practical efficiency.

\end{document}